\documentclass[conference]{IEEEtran}
\usepackage{cite}

\ifCLASSINFOpdf
   \usepackage[pdftex]{graphicx}
\else
\fi
\usepackage[cmex10]{amsmath}

\usepackage{mdwmath}
\usepackage{mdwtab}
\usepackage{fixltx2e}
\usepackage[justification=centering]{caption}

\usepackage{listings}
\usepackage{color}

\definecolor{dkgreen}{rgb}{0,0.6,0}
\definecolor{gray}{rgb}{0.5,0.5,0.5}
\definecolor{mauve}{rgb}{0.58,0,0.82}

\begin{document}
%
\title{Ensemble Framework for Real-time Decision Making}

\author{\IEEEauthorblockN{Philip Rodgers and John Levine}
\IEEEauthorblockA{Department of Computer and Information Sciences\\University of Strathclyde\\
Glasgow, UK\\
Email: \{philip.rodgers, john.levine\}@strath.ac.uk}}


%


\maketitle

\begin{abstract}

This paper introduces a new framework for real-time decision making in video games.  An Ensemble agent is a compound agent composed of multiple agents, each with its own tasks or goals to achieve.  Usually when dealing with real-time decision making, reactive agents are used; that is agents that return a decision based on the current state.  While reactive agents are very fast, most games require more than just a rule-based agent to achieve good results.  Deliberative agents---agents that use a forward model to search future states---are very useful in games with no hard time limit, such as Go or Backgammon, but generally take too long for real-time games.  The Ensemble framework addresses this issue by allowing the agent to be both deliberative and reactive at the same time.  This is achieved by breaking up the game-play into logical roles and having highly focused components for each role, with each component disregarding anything outwith its own role.  Reactive agents can be used where a reactive agent is suited to the role, and where a deliberative approach is required, branching is kept to a minimum by the removal of all extraneous factors, enabling an informed decision to be made within a much smaller time-frame.  An Arbiter is used to combine the component results, allowing high performing agents to be created from simple, efficient components.
\end{abstract}


%
\IEEEpeerreviewmaketitle

\section{Introduction}

\subsection{Ensemble Systems}
Ensemble based systems have been used for classification problems since the late 1970s \cite{1455590}, partitioning the features and using multiple classifiers.  A modern example of a powerful ensemble system is IBM's \emph{Watson}\cite{watson}.  Watson was originally created to play the TV quiz show \emph{Jeopardy}, but has since been opened up for general use.  It uses natural language recognition to analyse questions and generate queries.  It then sends the queries to multiple sources of answers--known as \emph{many experts}---and combining the answers, calculating confidence levels for each of the answers.  An important thing to note is that adding a new answer source does not affect the other sources or the logic of the system, it simply adds another answer to the set of all answers.

\subsection{Real-time Ensemble Agents}
Most AI for decision making is based on some form of \emph{`if this, then that'} model, using finite-state machines or decision trees.  The idea behind the Ensemble framework is to take the concept of feature partitioning from classification systems and apply it to real-time games.  The Ensemble framework is used to build complex agents out of simple components, or \emph{voices}, each with a simple goal or agenda.  At the heart of the framework is an \emph{Arbiter} which takes the outputs, or \emph{opinions}, of the voices and generates the final decision.  The concept sounds similar to that of a \emph{subsumption architecture}, but it differs significantly in that the decision making is not being deferred from one component to the next.  Instead, all of the voices have an opinion all of the time and each voice contributes to the end result, even if only slightly.  The Ensemble framework is far more akin to the classification ensemble systems than subsumption architectures.

The basic idea behind the Ensemble framework is to take simple, efficient and highly focused AI components and combine them to create complex behaviour.  This framework allows AI components to be easily added, removed or replaced without altering the behaviour of the other components or requiring the alteration of the Ensemble algorithm.  It also allows for a set of general purpose, reusable AI components to be created and used across multiple domains.  For example, Monte-Carlo Tree-Search \cite{chaslot2008monte}.

\begin{figure}[]
 	\begin{center}
		\includegraphics[scale=0.8]{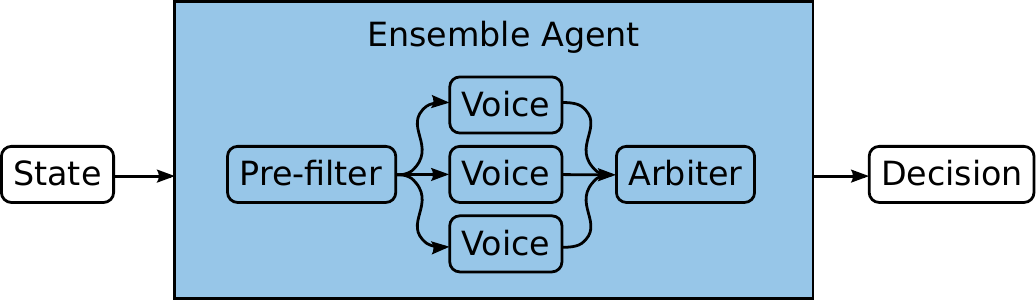}
		\centering
		\caption{Ensemble Agent}
		\label{fig:ensemble}
	\end{center}
\end{figure}

The Ensemble algorithm takes in the current game state, pre-filters the possible moves, sends this information to each of the component voices and combines the opinions of each voice into a single decision.  See figure \ref{fig:ensemble}.  The pre-filtering of moves allows for greater efficiency of the voices by removing any moves known to be invalid or known to be bad from previous iterations.

In its simplest form, the Ensemble was envisioned to have three primary components, with short-, middle- and long-range goals.  These can be seen as survival, tactics and strategy, respectively.  This idea is by no means a requirement, and the Ms. Pac-Man agent described in this paper does not strictly adhere to this structure.

For example, if the framework was to be used in a racing simulator, the components could be following the racing line, overtaking, defending your position and avoiding obstacles.

\subsection{Pac-Man}
The original Pac-Man is an arcade machine from 1980, created by game designer Toru Iwatani of Namco and considered one of the first examples of what would later become known as the \emph{survival horror} genre.  The game was hugely popular, and it was released in the United States in October 1980 by Midway Manufacturing Corporation of Illinois, USA.  

In Pac-Man, the player controls the main character around a maze using a four-way joystick.  The aim of the game is to complete each level by eating all the pills dotted around the maze whist avoiding the four antagonistic ghosts.  Each level also has four \emph{power-pills} that allow the main character to become \emph{energised} for a short period of time, during which the ghosts can be eaten.  Eating a power-pill also has the effect of making the ghosts reverse direction.  In later levels the time that the ghosts are edible drops to zero, so the reversal is the only effect of eating a power-pill.  Two bonus items also appear per level.

Points are scored by eating normal pills (10 points each), power pills (50 points each), ghosts (200, 400, 800 and 1600 points if eaten in succession) and bonuses (100, 300, 500, 700, 1000, 2000, 3000 or 5000, depending on the bonus).  From level 13 onwards, the bonus is always worth 5000 points.

\subsection{Ms. Pac-Man}
Ms. Pac-Man is a sequel to Pac-Man.  It has the same game mechanics but with several enhancements over the original game.  There are four mazes, as opposed to the single maze of Pac-Man, and the bonuses now move around the maze, rather than appearing stationary in the centre of the maze, and in the later levels the bonus is random.  The main difference, and what makes Ms. Pac-Man more appealing to players and AI developers alike, is the behaviour of the ghosts.

The original Pac-Man game is entirely deterministic, so players can learn patterns to complete each level.  The game can be beaten over and over by the simple repetition of the correct pattern.  Ms. Pac-Man introduced enough random ghost behaviour to allow for general strategies but not patterns.

Ms. Pac-Man was chosen for this project as it is well known and there is a lot of prior work.  Most of the prior work has been done using either the screen-capture Ms. Pac-Man competition framework\cite{screencapture}, or the Ms. Pac-Man vs. Ghost-Team framework.  This project uses our own emulator, written in Java and capable of playing the original Ms. Pac-Man code.  The emulator is described in section \ref{james}.

The original game has more complexity than the Ms. Pac-Man vs. Ghost-Team framework, and while the screen-capture framework is true to the original game, it has the computational overhead of converting the visible screen into usable information.  The use of an emulator enabled the agent to play the original game using an API that maps directly onto the emulator's RAM.

Ms. Pac-Man is a good benchmark of an AI system as it combines simplicity with high difficulty.  There are at most four possible options to choose from for any given state and the search space is confined to a single-screen maze.  Despite this simplicity, Ms. Pac-Man remains a hard problem for AI agents.  This is primarily because of the enclosed nature of the game space and the four-to-one ghost ratio.  Simply trying to keep a certain distance from the ghosts is likely to end up with Ms. Pac-Man being trapped. Understanding how the ghosts will react to a particular move, and so avoiding being trapped, is the key to survival in Ms. Pac-Man.

Ms. Pac-Man is not as simple as it first appears, mostly due the relative speeds and positions of the agents.  In the early levels, Ms. Pac-Man is approximately 25\% faster than the ghosts.  This speed ratio changes as the game progresses through the levels until level 21, where Ms. Pac-Man is approximately 25\% slower that the ghosts.  Whenever Ms. Pac-Man eats a pill she pauses for a single frame, so paths with pills are slower to traverse than clear paths, potentially allowing a ghost to catch up.  Ms. Pac-Man corners much faster than the ghosts; the ghosts always travel through the centre of each tile but Ms. Pac-Man has the ability to travel diagonally across corners, halving the distance travelled through that tile.  Three of the ghosts make their decisions based partly on which direction Ms. Pac-Man is facing, so a change in direction can have a significant effect on the behaviour of the ghosts.  When a ghost is eaten, all agents freeze momentarily, except the dead ghost eyes returning to base.  All these details must be taken into consideration when playing the game.

\subsection{High Scores}

The highest published score for an AI playing the original Ms. Pac-Man is 44,630 \cite{6374167}.  This score was the maximum of 100 games, with level six being the highest reached.  This would be considered a good score for a human, but the best human players can reach scores in excess of 900,000, clearing more than 130 levels\cite{twin}.  The highest score recorded by the Ensemble agent is currently 162,280 at level 24, but this result was achieved while recording a video and is not part of the experimental data\cite{youtube}.

Ms. Pac-Man was recently released on Steam.  The leader board would suggest 30,000 to be a reasonable average score for a human.  In discussion with Patrick Scott Patterson---a video game advocate, journalist and record holder---Mr. Patterson suggested that six-figure scores were rare, and that only a handful of players in the world are capable of playing the game at this level.  The current human world record is 933,580, set by Abdner Ashman in 2006.  Only five people have officially reached over 900,000 points.

\section{Ms. Pac-Man vs. Ghost-Team}
Initial experiments for this project were done using the Ms. Pac-Man vs. Ghost-Team framework.  This led to some useful insights and a very capable agent---usually reaching the global time-limit at around level 12, often without losing a life.  The agent played especially well against highly predictable ghosts, such as the aggressive ghost team, where it could group the ghosts together, eat a power-pill and then eat all the ghosts in quick succession.

To find out how the agent would fare against a top-ranking ghost team, we contacted the author of the \emph{Memetix} ghost team, Daryl Tose, who very graciously sent us his code.  As it turned out, despite the \emph{Memetix} ghost team being almost completely deterministic, the Ensemble agent rarely got past the first level.  The conclusion being that however strong the Ms. Pac-Man agent is, the ghosts can always win if they work as a team.

Unfortunately, the Ms. Pac-Man vs. Ghost-Team competition is no longer running and the website is down.  We were unable to see how the agent compared to other competition agents. 

\section{James} \label{james}
James was written from the ground up to be an object-oriented Ms. Pac-Man emulator.  Large sections of the code came from the ArcadeFlex project\cite{arcadeflex}\cite{arcadeflexcode}.  The code was constructed as a core emulator, with a full emulator built around it.  The core emulator emulates the CPU, RAM and I/O.  The full emulator adds windowing, graphics and keyboard support.  This allows the core emulator to be used as a forward model not tied to the 60 frames per second of the full emulator.

An API for agents to interact with the game was created.  Game state information is obtained by interpreting the contents of specific memory locations within the emulator.  Actions are performed by setting the values of the memory-mapped I/O ports.  A graph data structure was also created for maze-based queries.  All-pairs tile distances were pre-calculated using the Floyd-Warshall algorithm\cite{warshall}.  In addition to the all-pairs distances, every tile stores the distance to every other tile for each available move.  These directional distances were pre-calculated using A* search with the true minimum distance as the heuristic.

Using the emulated Ms. Pac-Man code as a forward model is extremely accurate\footnote{Too accurate, in fact.  The pseudo-random number generator has to be re-seeded in order to make the forward model non-deterministic.}, but very inefficient.  To counter this, an alternative forward model---the \emph{simulator}---was created.  The simulator is a native Java partial model of the game.  A lot of work went into making the simulator as accurate as possible, especially with regard to ghost behaviour.  Although the simulator it is not 100\% accurate, it is generally accurate enough if synchronised with the emulator before each use.  The simulator is very fast compared to the emulator, more than making up for the loss of accuracy.

\section{Ensemble Agent for Ms. Pac-Man}

For the Ms. Pac-Man agent, the tasks were defined as:
\begin{itemize}
\item Eat pills.  This scores a modest amount of points, but its primary purpose is to finish the level.  This is both a short and long range goal.  Eating the next pill is a short range goal, eating \emph{all} the pills is a long range goal.
\item Eat fruit.  This scores quite a lot of extra points, depending on the fruit, but is not essential.  Medium range goal.
\item Eat ghosts.  This is a great way to score extra points during the early levels, but becomes impossible in the later levels.  Medium range goal.
\item Avoid ghosts.  Staying alive is, obviously, of primary concern.  The ghosts are the only adversarial agents in the game, so knowing how not to get caught is key to survival.  Short to medium range goal.
\end{itemize}

\begin{figure}[]
 	\begin{center}
		\includegraphics[scale=1.0]{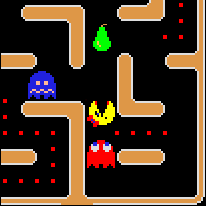}
		\centering
		\caption{Opposing opinions}
		\label{fig:options}
	\end{center}
\end{figure}

It became apparent during initial experiments that simply having each voice offering its preferred move at any given point lead to a lot of deadlocks.  The voices would often have opposing opinions due to the completely disparate nature of their goals.  In figure \ref{fig:options}, Ms. Pac-Man is approaching a junction with three options: UP, LEFT or DOWN.  The pill eating voice will vote to go DOWN, the ghost eating voice will vote to go LEFT and the fruit eating voice will vote to go UP.  No reward is worth dying for, so the ghost avoiding voice will veto DOWN, leaving a deadlock between UP and LEFT.  The move could be picked at random, or the arbiter could be crafted with some domain knowledge to make a more informed decision, but the Ensemble framework allows the use of \emph{fuzzy logic} where the voices \emph{rate} each available move according to its goal.

Using the same scenario as in Figure \ref{fig:options}, with DOWN vetoed by the ghost avoiding voice, the other three voices need to present their ratings for UP and LEFT.  As the voices are all distance based, the ratings will be the inverse of the distance to the goal in each available direction.  If the voices are weighted equally, the resulting move would be UP.  An approximation of the calculation can be seen in Table \ref{tab:values}.

\begin{table}[h]
\renewcommand{\arraystretch}{1.4}
\begin{center}
\begin{tabular}{ |c|c|c|c|c| } 
 \hline
  & Pill eater & Fruit eater & Ghost eater & Sum (approx.) \\ 
\hline
 UP & 1/7 & 1/3 & 1/24 & 1/2 \\ 
 \hline
 LEFT & 1/7 & 1/24 & 1/4 & 2/5\\ 
 \hline
\end{tabular}
\end{center}
\caption{Calculating move values}
\label{tab:values}
\end{table}

This solution is far less likely to lead to a tie-break situation, and it is also more flexible in terms of weighting each voice.  In the above example the agent decided to go UP, essentially because the fruit is closer than the edible ghost.  But it was a close call.  Generally, there is likely to be more chance of eating the fruit in the future than the ghost, so LEFT would probably have been a better choice.  Weighting the ghost eater higher than the fruit eater would have changed the decision to LEFT.  Rating the pill eater low---because pills are low value and static---would likely make the agent head towards the fruit after eating the ghost, and so a powerful strategy is emerging from the simple rules.  The arbiter never actually targets anything, or makes any sort of plan, it simply chooses the highest combined-rated move at any given point.

The final Ensemble agent for Ms. Pac-Man in \emph{James} is composed of four voices, with an arbiter taking the opinions of each voice and combining them to make the final decision.
\begin{itemize}
\item Ghost Dodger.  Avoiding ghosts is the most important aspect of the game, and is also the hardest to do, computationally.  This voice is discussed in detail later.
\item Pill Muncher.  This voice rates each move as the inverse of the tile distance to the nearest pill in that direction.  Pills near ghosts are artificially made to look further away, meaning that the pill muncher will rate safe pills higher than those with ghosts near by.
\item Fruit Muncher.  This voice has no opinion unless there is a fruit bonus on the screen.  If a fruit is on the screen, the voice will attempt to intercept it.  It rates the available moves as the inverse of the tile distance to the fruit.
\item Ghost Muncher.  This voice only has an opinion if Ms. Pac-Man is energised.  It rates each move based on the distance to the nearest edible ghost in that direction.  It also tries to avoid eating a power-pill until all ghosts are out of their base.
\end{itemize}


\subsection{Arbitration}

The arbiter holds weights for each voice, and the rating for each move is calculated as the sum of each voice, not including Ghost Dodger, multiplied by its weight. This sum is then multiplied by Ghost Dodger’s rating, which is also multiplied by its weight for normalising the overall rating, to give the final rating for the move.

$$R_{m}=(\sum_{n=2}^{i=n}V_{i,m}W_{i})V_{1,m}$$

$R$ is a vector of move ratings, $V$ is a matrix of voice move ratings where $V_{1}$ always the Ghost Dodger, $W$ is a vector of voice weights, $m$ is the move being rated and $n$ is the total number of voices. The arbiter will choose the move corresponding to the highest value in $R$. If more than one move has the highest rating, the arbiter will select at random from the highest rated moves.

As a proof of concept, a simple one-plus-one evolutionary strategy was used to optimise the weights of the various voices.  A baseline score was recorded over 100 games, then the weights were adjusted by a small random amount.  Another 100 games were played with the new weights, with the new weights being kept if the average score improved.  This was done 1000 times, for a total of 100,000 games. An early version of the ensemble AI was used for this experiment, and it showed the ability to increase the ability of that player.  In this experiment, a modest but noticeable increase in average score of approximately 30\% was achieved.  Unfortunately this technique takes a very long time, and it was not used in any of the final agents.

Dynamic weighting is also a possibility, combining the Ensemble arbiter with a finite-state machine.  Using such a technique would allow the agent to adjust the voice weights depending on the situation.  This could be of use in, for example, stealth games.  One set of weights could be used during stealthy missions or portions of missions, with another set of weights used in situations where the player's stealth has been compromised.  Dynamic weighting could also be used in general video games playing (GVGP)\cite{levine2013general}, especially when combined with general-purpose components.

\subsection{Ghost Dodging}

The final Ghost Dodger algorithm uses the simulator for depth limited search, but rates each move based on random sampling.  This algorithm is closely related to the averaged depth-limited search technique applied to the game 2048, as demonstrated at the IEEE CIG2014 conference in Dortmund\cite{2048}.

The algorithm is given 10ms in which to make random depth-limited samples through the maze.  Every time it reaches its depth limit of 8 without dying, the initial move's score gets incremented.  At the end of the 10ms, the voice returns its rating for each move, based on how many times it reached the depth limit.  The depth of 8 was chosen as a trade-off between depth and the number of samples.  For this algorithm, a \emph{move} is a straight line from the current position to the next corner or junction.  This simplifies the algorithm as Ms. Pac-Man's direction does not need to be re-calculated mid-move for cornering.

\section{Monte-Carlo Tree-Search}
The Monte-Carlo Tree-Search (MCTS) based agent was created to set a benchmark for the Ensemble agent.  This agent sets a target tile in the maze that is the next decision point\footnote{The initial `bootstrap' target is (14,24) with the move set as LEFT.  The tile (15,24) is always Ms. Pac-Man's starting position, and LEFT to (14,24) is a valid safe move in all four mazes.} in the current direction.  The MCTS algorithm then utilises the time it takes to reach the target to calculate what the best move and new target will be when it gets there, thus giving the algorithm hundreds of milliseconds to deliberate on each move.

The searching is done using the simulator, with the nodes in the tree storing the move needed to advance the simulation to that point.  Initially, a full snapshot of the game-state was stored in the nodes, with the simulator being synchronised to that snapshot each time the root node has the selectAction() method called.  Storing just the move turned out to be more efficient, and also more accurate when the ghost movements are random, than storing whole snapshots at the nodes.

The simulator includes a scoring system, and score deltas are used in place of roll-outs.  Because of this, the agent is not strictly MCTS.  Using a standard roll-out meant playing to either death or a win.  Wins are easy to reward, but how do you reward death, given that most roll-outs end in death?  A standard MCTS player with roll-outs was tested, but it performed badly, even when using a depth-limited roll-out.

It is worth pointing out that the MCTS agent ignores the fruit bonuses as they are not modelled in the simulator.  This causes the agent to miss out on a lot of extra points.  Another experiment was run using the Ensemble agent with the fruit munching voice disabled for more comparable results.

Using the UCT algorithm unmodified lead to the MCTS creating highly asymmetric trees, so the UCT algorithm was adjusted to create more symmetric trees.  Experiment runs were made using both configurations.

\section{Experiments}
For the experiments, each player played 100 full games.  With the exception of the one-tenth speed experiment with the emulator, all games were run at normal game speed.  The experiments run were:

\begin{itemize}
\item Hello World AI.  This is a purely reactive agent based on the Hello World agent from the Ms. Pac-Man vs. Ghost Team framework
\item Ensemble agent using the emulator and the safe-path ghost avoidance algorithm to a depth of three
\item Ensemble agent using the emulator and the safe-path ghost avoidance algorithm to a depth of three at one-tenth speed
\item Ensemble agent using the simulator and the safe-path ghost avoidance algorithm to a depth of three
\item Ensemble agent using the simulator and the safe-path ghost avoidance algorithm to a depth of eight
\item Ensemble agent using the simulator and the safe-path ghost avoidance algorithm to a depth of twenty
\item MCTS agent with the UCT formula adjusted for symmetric trees
\item MCTS agent with the UCT formula adjusted for asymmetric trees
\item Final Ensemble agent with the fruit munching component removed to better match the MCTS agents
\item Final Ensemble agent
\item Final Ensemble agent with the ability to use the emulator removed
\end{itemize}

\subsection{Hello World AI}
\begin{figure}[]
 	\begin{center}
		\includegraphics[scale=0.6]{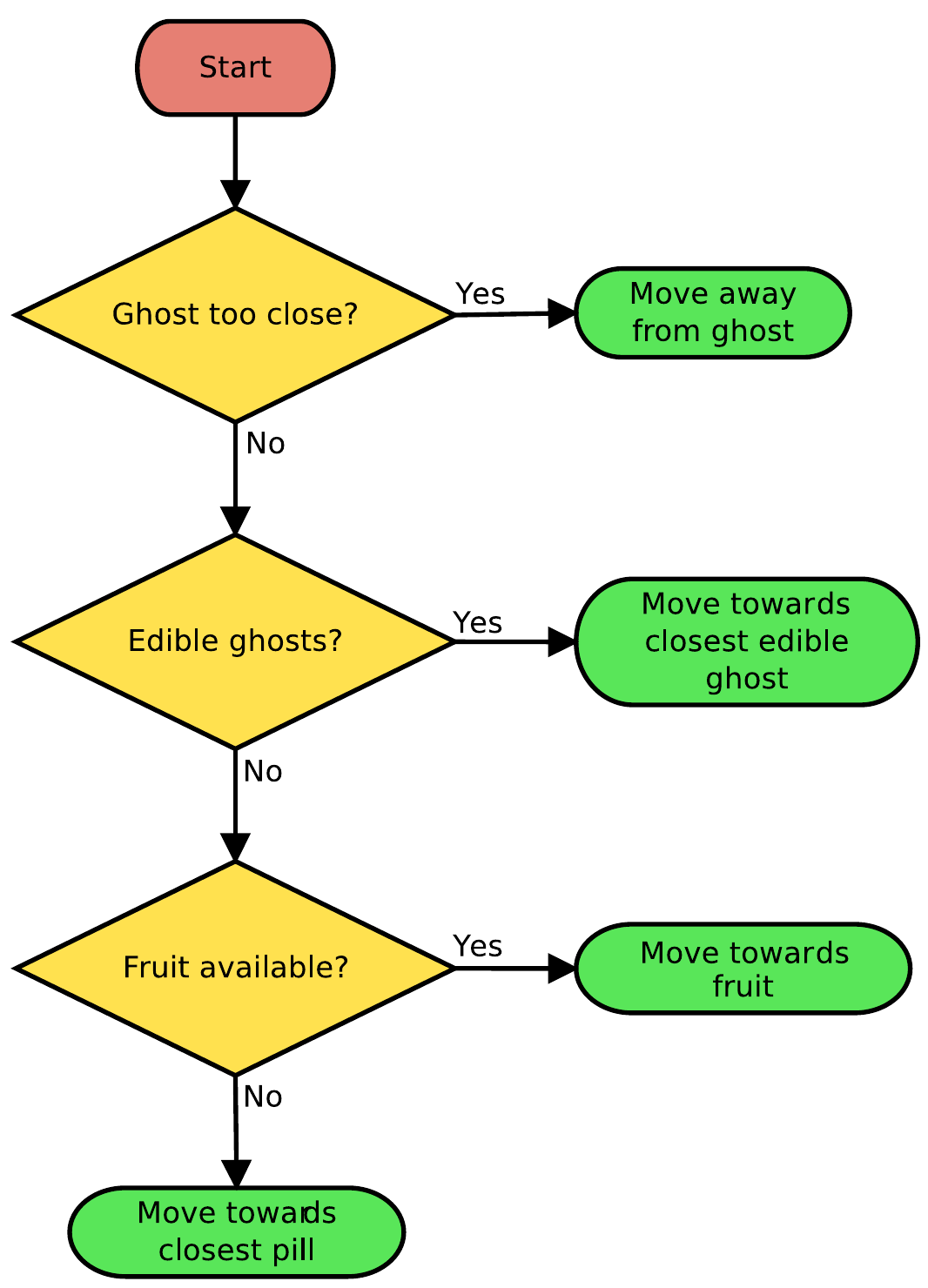}
		\centering
		\caption{Hello World AI Behaviour Tree}
		\label{fig:hello}
	\end{center}
\end{figure}

The Hello World agent is a reactive agent with a simple behaviour tree (see figure \ref{fig:hello}).  It is based off the Hello World agent from the Ms. Pac-Man vs. Ghost-Team framework.  It covers all four key behaviours required for playing Ms. Pac-Man, but it does so in order of precedence.  The ghost avoidance is weaker than that of the other agents, as it only knows where the ghosts are, not where they are going to be.

\subsection{Single Safe Path}
The first Ghost Dodger technique used the core emulator as a forward model.  Running the full game on emulated hardware is very inefficient, but extremely accurate.  This produced an agent that was short sighted, but very good at exploiting the so called \emph{walk-through bug} when trapped in a corner\footnote{The walk-through bug---also exploited by the best human players---is the result of how the original game detects collisions.  Ms. Pac-Man and a ghost only collide if their centre points are in the same 8x8 pixel tile at the same time.  If a ghost is in tile X, and Ms. Pac-Man is in tile Y in one frame, and in the next frame Ms. Pac-Man is in tile X and the ghost in tile Y, no collision is detected and Ms. Pac-Man is free to continue on her way.}.  The emulator forward model is too inefficient to use as a means of rating each move, so instead it was used to veto any moves leading to a quick death.

The algorithm used had to be very efficient because the emulator is not.  The algorithm is a depth-limited search of just three moves, where a move is travelling from one decision point to the next.  A decision point is either a junction in the maze, a power pill or an artificial decision point used to break up a path that would otherwise be too long.  For speed, the algorithm both fails fast and succeeds fast.  As soon as a single safe path of three moves is found, the algorithm returns \emph{true}.

For greater efficiency, the simulator was used in place of the emulator.  Over 100 games, the simulator-based agent did \emph{better} than the `perfect' emulator-based agent.  The reason for this would appear to be timing.  The game and the AI run asynchronously in separate threads, with the game advancing at 60 frames per second.  If the AI agent takes longer than 16ms to make its decision, the game state will have changed; not by much, but possibly significantly.  To account for this, and to set a benchmark, we ran another experiment of 100 games using the emulator-based agent but with the game slowed down to one-tenth speed, giving the agent 160ms between frames.

Because the simulator is so much faster than the emulator, we ran experiments using the same algorithm but searching to a depth of 8 and to a depth of 20, both of which returned results well within 16ms.  While searching deeper did show an improvement, it was negligible.

\subsection{MCTS}
The MCTS player was developed to set a high bar and demonstrate a purely deliberative agent.  Despite using only the simulator and hence no knowledge of fruit, the MCTS players managed to play to a high standard.  The asymmetric MCTS did a much better job of surviving than the more symmetric version.  Both, however, scored roughly the same.  This would suggest that the symmetric version was better at scoring points, but more likely to get killed doing so.

We are confident that the MCTS player is a reasonable benchmark and a powerful example of a deliberative agent

\subsection{Final Ensemble}
Along with the final Ensemble agent, two variants were also used in the experiments.  The `no fruit' version was created for a fairer comparison with the MCTS agent.  Not trying to get the fruit should, in theory, make the agent score fewer points per level but survive longer as the agent will not be lured into dangerous situations by the prospect of eating fruit.

The `simulator only' version was created to see what effect not having access to the emulator would have.  The final Ensemble agent uses the emulator as an extremely short range forward model to see the immediate outcome of the current move.  Although this look-ahead is only eight frames, it allows the agent to better handle close-quarters interaction with the ghosts.  The simulator is good, but not perfect.  Being out by a couple of pixels makes little difference at long range, but it can make all the difference at close range.

\section{Results}
\begin{table}
\renewcommand{\arraystretch}{1.4}
\begin{center}
\begin{tabular}{ |c|c|c|c|c| } 
 \hline
 Player & Mean & Min. & Max. & Std. Dev. \\ 
\hline
 Hello World AI & 12735 & 3290 & 32780 & 6759 \\ 
 \hline
 Ensemble (Emu 3 safe) & 39297 & 4540 & 68910 & 14943\\ 
 \hline
 Ensemble (Emu 3 slow) & 71259 & 9360 & 136310 & 26383\\ 
 \hline
 Ensemble (Sim 3 safe) & 46672 & 6250 & 109190 & 21528\\ 
 \hline
 Ensemble (Sim 8 safe) & 51447 & 13600 & 105440 & 19175\\ 
 \hline
 Ensemble (Sim 20 safe) & 55843 & 15350 & 101660 & 21258\\ 
 \hline
 MCTS (symmetric) & 57151 & 5260 & 95560 & 17068\\ 
 \hline
 MCTS (asymmetric) & 58058 & 19500 & 96120 & 16155\\ 
 \hline
 No Fruit Ensemble & 69779 & 21990 & 105770 & 22236\\ 
 \hline
 Final Ensemble & 102238 & 8060 & 155640 & 29520\\ 
 \hline
 Sim Only Ensemble& 74820 & 9160 & 124440 & 25573\\ 
 \hline
\end{tabular}
\end{center}
\caption{Agent scores over 100 games}
\label{tab:scores}
\end{table}
\begin{table}
\renewcommand{\arraystretch}{1.4}
\begin{center}
\begin{tabular}{ |c|c|c|c|c|c| } 
 \hline
 Player & Mean & Min. & Max. & Mode & Median \\ 
\hline
 Hello World AI & 2.09 & 1 & 5 & 2 & 2 \\ 
 \hline
 Ensemble (Emu 3 safe) & 5.83 & 1 & 10 & 6 & 6 \\ 
 \hline
 Ensemble (Emu 3 slow) & 9.76 & 2 & 21 & 7 & 9 \\ 
 \hline
 Ensemble (Sim 3 safe) & 6.67 & 2 & 15 & 6 & 6 \\ 
 \hline
 Ensemble (Sim 8 safe) & 7.4 & 3 & 14 & 8 & 7 \\ 
 \hline
 Ensemble (Sim 20 safe) & 8.04 & 3 & 16 & 8 & 8 \\ 
 \hline
 MCTS (symmetric) & 9.83 & 1 & 17 & 9 & 9 \\ 
 \hline
 MCTS (asymmetric) & 15.21 & 6 & 22 & 15 & 15.5 \\ 
 \hline
 No Fruit Ensemble& 14.55 & 4 & 23 & 21 & 15 \\
 \hline
 Final Ensemble & 15.68 & 1 & 22 & 21 & 16.5 \\ 
 \hline
 Sim only Ensemble& 11.21 & 2 & 22 & 9 & 10 \\ 
 \hline
\end{tabular}
\end{center}
\caption{Levels reached over 100 games}
\label{tab:levels}
\end{table}

\begin{figure}[]
 	\begin{center}
		\includegraphics[scale=0.34, trim=2cm 3cm 2cm 3cm]{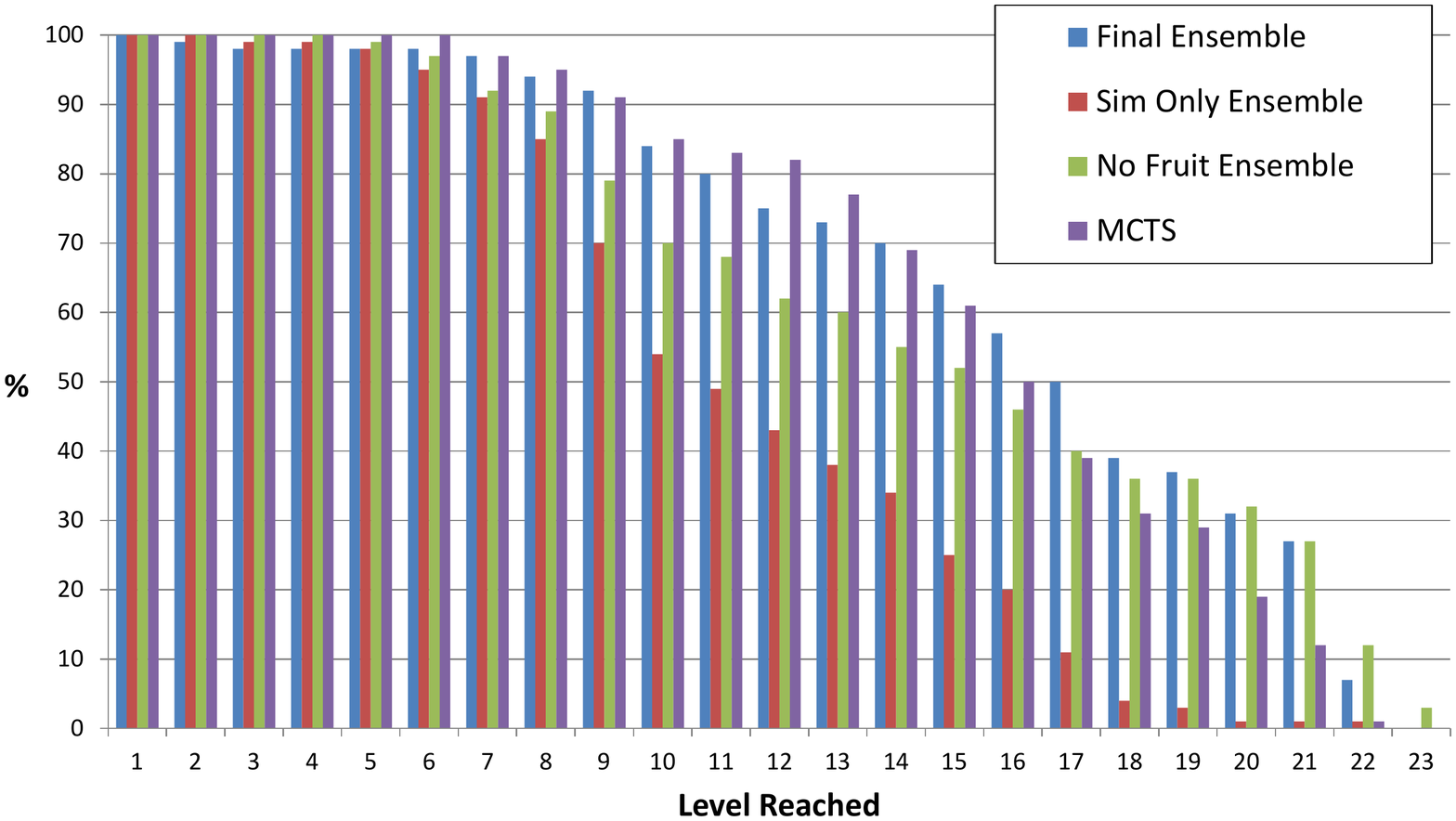}
		\centering
		\caption{Survival rates}
		\label{fig:survival}
	\end{center}
\end{figure}

\begin{figure}[]
 	\begin{center}
		\includegraphics[scale=0.5, trim=2cm 9cm 2cm 11cm]{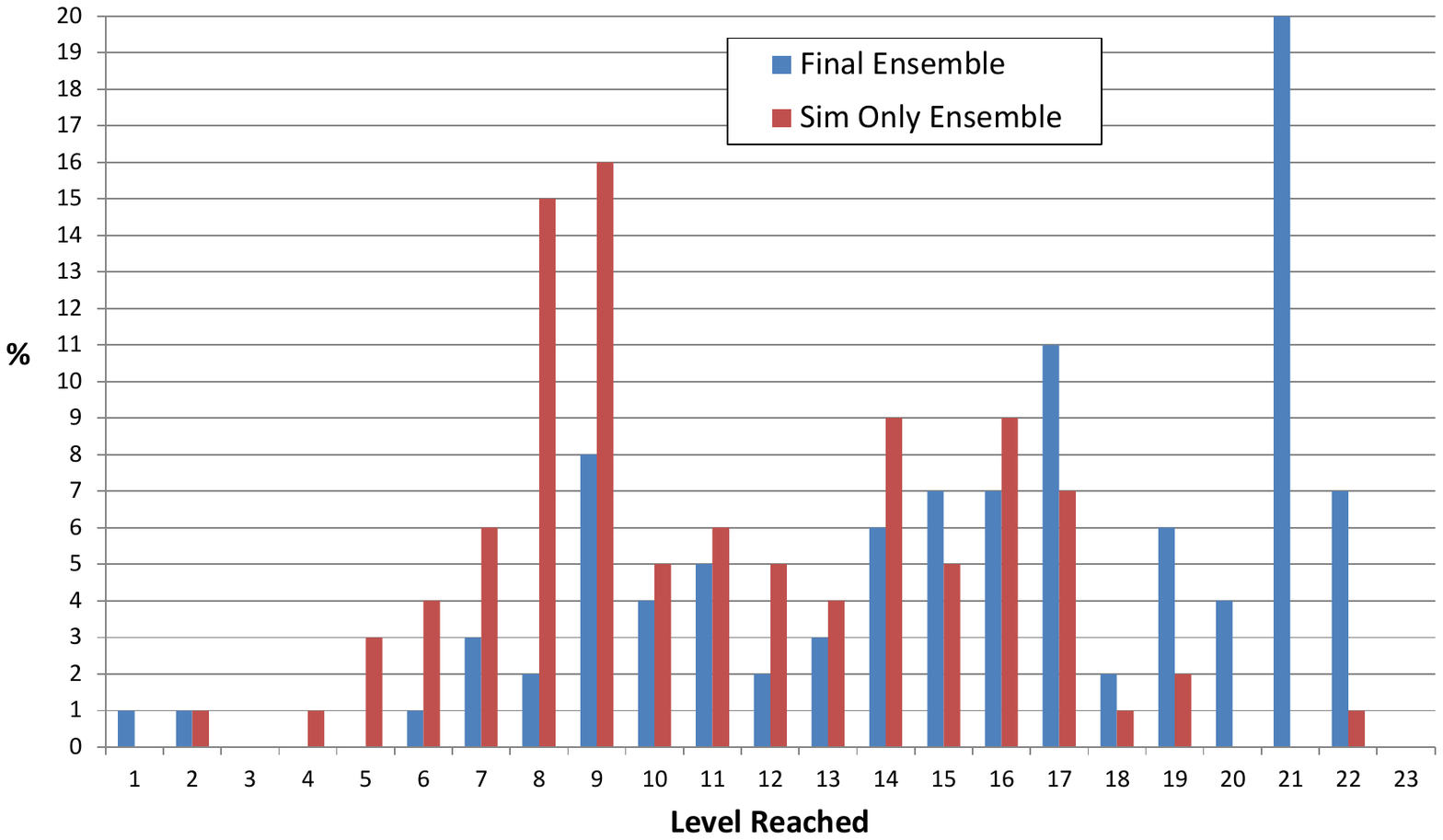}
		\centering
		\caption{With and without access to the emulator}
		\label{fig:emu-sim}
	\end{center}
\end{figure}

\begin{figure}[]
 	\begin{center}
		\includegraphics[scale=0.5, trim=2cm 9cm 2cm 11cm]{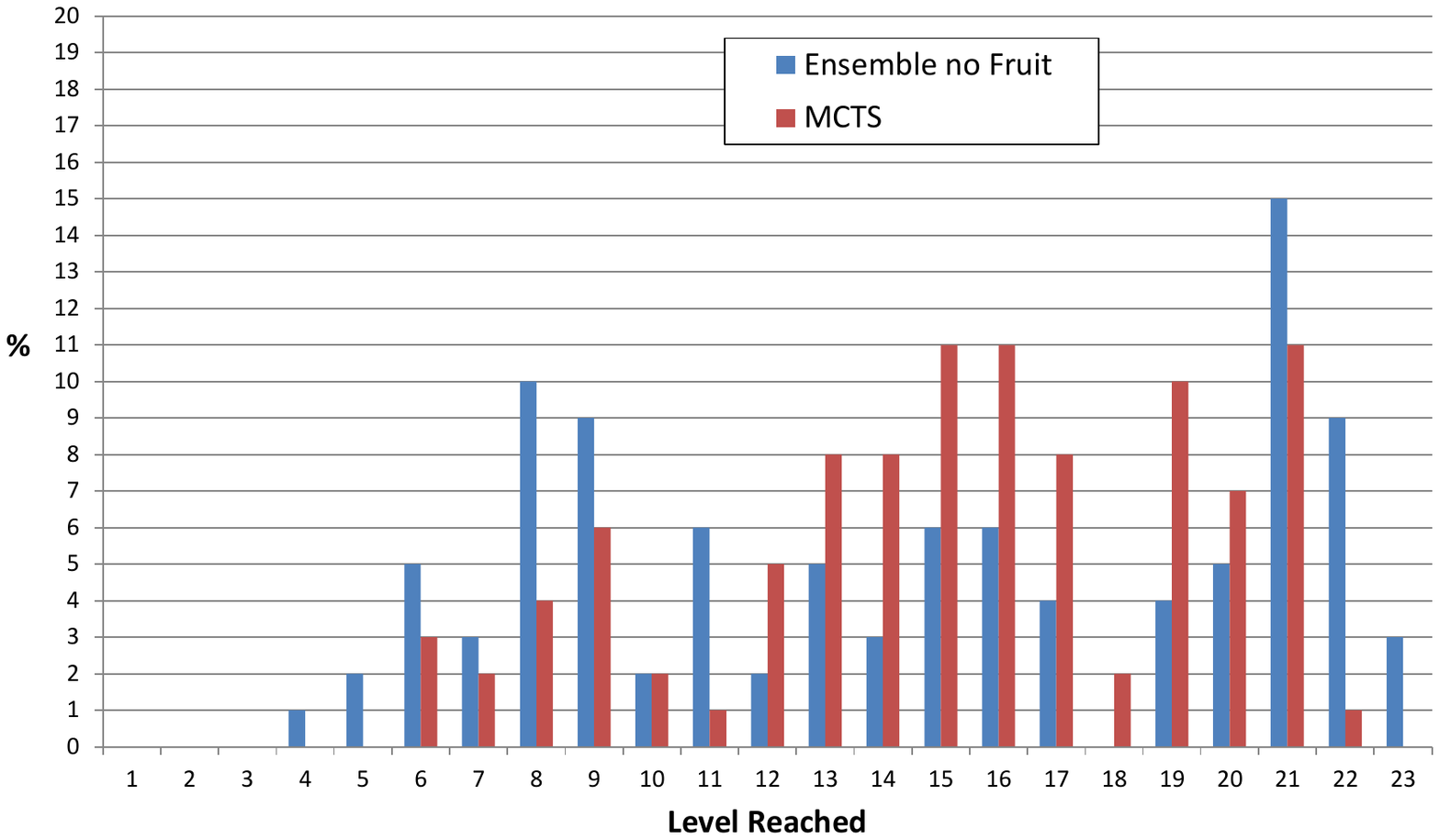}
		\centering
		\caption{Ensemble (no fruit) vs. MCTS}
		\label{fig:no-fruit}
	\end{center}
\end{figure}

Table \ref{tab:scores} shows the results of the experiments in terms of points scored.  Table \ref{tab:levels} shows the results in terms of levels reached.  Figure \ref{fig:survival} shows the relative survivability of the four top agents.  It shows the percentage that an agent reached a particular level during the experiment.

Although not a clean sweep for the final Ensemble, it is the clear winner.  It has much higher mean and maximum scores, it has the highest mean, mode and median in terms of level reached and is only beaten by the `no fruit' Ensemble for highest level reached.

A comparison of level reached between final Ensemble and the simulator only Ensemble can be seen in figure \ref{fig:emu-sim}.  This chart shows that emulator does provide a powerful tool to the Ensemble in terms of level reached.  There is a peak at levels 8 and 9 that is far greater for the `simulator only' Ensemble.  The emulator helps the final Ensemble through these early levels on its way to a large peak at level 21.  Although the sim-only Ensemble has a very respectable high score and highest level, the averages are significantly lower.

The `no fruit' variant performed mostly as expected.  It did not score as well as the final Ensemble, but it did get higher minimum and maximum levels reached.  Figure \ref{fig:no-fruit} shows a comparison of levels reached for the `no fruit' Ensemble and the MCTS agent.  From this we can see that the MCTS agent coped better during the earlier levels, but the `no fruit' Ensemble fared better in the later levels.  Overall, the `no fruit' Ensemble agent scored modestly better than the MCTS agent.  It also has a much higher mode level of 21, compared to 15 for the MCTS agent, but the deaths in the early levels brings the mean level reached to just below that of the MCTS agent.

\section{Conclusion}
The results in this paper show that, for Ms. Pac-Man at least, the Ensemble framework can be used to create real-time agents that perform as well as, if not better than, deliberative agents.  The separation of components that can be purely reactive from those requiring deliberation allows the deliberative components to be as efficient as possible.  Combining the fuzzy logic opinions of the component voices in a non-sequential manner does seem to generate deliberative level behaviour in a far more reactive time frame.

\section{Future Work}

The simple arbiter could be replaced by something more sophisticated and dynamic; possibly a trained neural network or a genetic algorithm to learn a strategic sense of the game.  The experiments evolving the voice weights of the ensemble were mildly successful, but slow and tedious.  It does, however, leave open the possibility of using deep reinforcement learning techniques to develop \emph{bigger picture} strategies.

The results for Ms. Pac-Man are better than anticipated, but it is a sample of one.  The Ensemble framework needs to be applied to more and varied games before it can can be fully justified as a general framework for real-time decision making.

\section{Code}

The code used in this project can be downloaded from GitHub at https://github.com/philrod1/james


\section*{Acknowledgements}

George Moralis and the ArcadeFlex developers.

Scott Lawrence for the Ms. Pac-Man disassembly.



\bibliographystyle{IEEEtran}
\bibliography{IEEEabrv,mybibfile}

\begin{thebibliography}{10}
\providecommand{\url}[1]{#1}
\csname url@samestyle\endcsname
\providecommand{\newblock}{\relax}
\providecommand{\bibinfo}[2]{#2}
\providecommand{\BIBentrySTDinterwordspacing}{\spaceskip=0pt\relax}
\providecommand{\BIBentryALTinterwordstretchfactor}{4}
\providecommand{\BIBentryALTinterwordspacing}{\spaceskip=\fontdimen2\font plus
\BIBentryALTinterwordstretchfactor\fontdimen3\font minus
  \fontdimen4\font\relax}
\providecommand{\BIBforeignlanguage}[2]{{%
\expandafter\ifx\csname l@#1\endcsname\relax
\typeout{** WARNING: IEEEtran.bst: No hyphenation pattern has been}%
\typeout{** loaded for the language `#1'. Using the pattern for}%
\typeout{** the default language instead.}%
\else
\language=\csname l@#1\endcsname
\fi
#2}}
\providecommand{\BIBdecl}{\relax}
\BIBdecl

\bibitem{1455590}
B.~Dasarathy and B.~V. Sheela, ``A composite classifier system design: Concepts
  and methodology,'' \emph{Proceedings of the IEEE}, vol.~67, no.~5, pp.
  708--713, May 1979.

\bibitem{watson}
D.~Ferrucci, E.~Brown, J.~Chu-Carroll, J.~Fan, D.~Gondek, A.~A. Kalyanpur,
  A.~Lally, J.~W. Murdock, E.~Nyberg, J.~Prager \emph{et~al.}, ``Building
  watson: An overview of the deepqa project,'' \emph{AI magazine}, vol.~31,
  no.~3, pp. 59--79, 2010.

\bibitem{chaslot2008monte}
G.~Chaslot, S.~Bakkes, I.~Szita, and P.~Spronck, ``Monte-carlo tree search: A
  new framework for game ai.'' in \emph{AIIDE}, 2008.

\bibitem{screencapture}
\BIBentryALTinterwordspacing
S.~Lucas. Ms pac-man competition. [Online]. Available:
  \url{http://csee.essex.ac.uk/staff/sml/pacman/PacManContest.html}
\BIBentrySTDinterwordspacing

\bibitem{6374167}
G.~Foderaro, A.~Swingler, and S.~Ferrari, ``A model-based cell decomposition
  approach to on-line pursuit-evasion path planning and the video game ms.
  pac-man,'' in \emph{2012 IEEE Conference on Computational Intelligence and
  Games (CIG)}, Sept 2012, pp. 281--287.

\bibitem{twin}
\BIBentryALTinterwordspacing
{Ms. Pac-Man top 10}. [Online]. Available:
  \url{http://www.twingalaxies.com/scores.php?scores=1386}
\BIBentrySTDinterwordspacing

\bibitem{youtube}
\BIBentryALTinterwordspacing
``{Ms. Pac-Man AI. New high score.}'' [Online]. Available:
  \url{https://youtu.be/Y9YazqWaEAM}
\BIBentrySTDinterwordspacing

\bibitem{arcadeflex}
\BIBentryALTinterwordspacing
G.~Moralis. Arcadeflex. [Online]. Available: \url{http://www.arcadeflex.com}
\BIBentrySTDinterwordspacing

\bibitem{arcadeflexcode}
\BIBentryALTinterwordspacing
------. Arcadeflex code. [Online]. Available:
  \url{https://github.com/georgemoralis/arcadeflex036}
\BIBentrySTDinterwordspacing

\bibitem{warshall}
\BIBentryALTinterwordspacing
S.~Warshall, ``A theorem on boolean matrices,'' \emph{J. ACM}, vol.~9, no.~1,
  pp. 11--12, Jan. 1962. [Online]. Available:
  \url{http://doi.acm.org/10.1145/321105.321107}
\BIBentrySTDinterwordspacing

\bibitem{levine2013general}
J.~Levine, C.~B. Congdon, M.~Ebner, G.~Kendall, S.~M. Lucas, R.~Miikkulainen,
  T.~Schaul, and T.~Thompson, ``General video game playing,'' \emph{Dagstuhl
  Follow-Ups}, vol.~6, 2013.

\bibitem{2048}
\BIBentryALTinterwordspacing
P.~Rodgers and J.~Levine, ``An investigation into 2048 {AI} strategies,'' in
  \emph{2014 {IEEE} Conference on Computational Intelligence and Games, {CIG}
  2014, Dortmund, Germany, August 26-29, 2014}.\hskip 1em plus 0.5em minus
  0.4em\relax {IEEE}, 2014, pp. 1--2. [Online]. Available:
  \url{http://dx.doi.org/10.1109/CIG.2014.6932920}
\BIBentrySTDinterwordspacing

\end{thebibliography}

\end{document}